# It is hard to see a needle in a haystack: Modeling contrast masking effect in a numerical observer

Ali R. N. Avanaki[1], Kathryn S. Espig[1], Albert Xthona[1], Tom R. L. Kimpe[2], Predrag R. Bakic[3], Andrew D. A. Maidment[3]

[1] Barco Healthcare, Beaverton, OR, USA
[2] BARCO N.V., Healthcare Division, Kortrijk, Belgium
[3] University of Pennsylvania, Department of Radiology, Philadelphia, PA, USA

**Abstract.** Within the framework of a virtual clinical trial for breast imaging, we aim to develop numerical observers that follow the same detection performance trends as those of a typical human observer. In our prior work, we showed that by including spatio-temporal contrast sensitivity function (stCSF) of human visual system (HVS) in a multi-slice channelized Hotelling observer (msCHO), we can correctly predict trends of a typical human observer performance with the viewing parameters of browsing speed, viewing distance and contrast. In this work we further improve our numerical observer by modeling contrast masking. After stCSF, contrast masking is the second most prominent property of HVS and it refers to the fact that the presence of one signal affects the visibility threshold for another signal. Our results indicate that the improved numerical observer better predicts changes in detection performance with background complexity.

**Keywords:** contrast masking effect, spatio-temporal contrast sensitivity function, channelized Hotelling observer, human visual system, virtual clinical trial, and psychometric function.

## 1       Purpose

Commonly used numerical observers cannot necessarily predict the behavior of a typical human observer in all observation scenarios. This is due to the fact that they are modeled after ideal observers (i.e., maximizing some detection performance metric) with some concessions for tractability (e.g., channelization). For example, a multi-slice channelized Hotelling observer (msCHO), without correct application of spatio-temporal contrast sensitivity function (stCSF), is unable to predict the fundamental effect of display contrast on detection of lesions in digital breast tomosynthesis (DBT) [1, 2].

Our goal is to enhance a numerical observer by making it perform more similarly to a human observer. Our approach towards this goal is to integrate important properties of the human visual system (HVS) as a pre-processing step to a commonly used numerical observer (msCHO). In other words, as the result of HVS modeling, "per-ceived" 3D image stacks are fed to msCHO. Previously we have reported on the benefits of integrating the HVS property of stCSF with msCHO [1, 2, 3].

In this work, we study the effect of modeling the contrast masking property of HVS in our numerical observer. Contrast masking refers to the phenomenon that the presence of a signal ("masker") makes detection of another signal ("maskee") more difficult.

### 1.1 Prior work

Zhang *et al* channelized the input image in orientation and (spatial) frequency [4]. In the most sophisticated model used, an inhibitory component (denominator of Eq. 13 therein) is used to factor in the contrast masking effect.

Early channelization of the data in [4] is undesirable in our methodology. By postponing channelization to CHO (i.e., HVS simulation followed by a traditional numerical observer at the backend), one can replace msCHO by a more sophisticated observer to upgrade the pipeline. In other words, early channelization discards data that may be useful to the detection task to be performed at the backend. Also, the perception model in [4] is parametric and is unusable without a calibration to (psychophysical) experimental results.

Krupinski *et al* developed a perceptual numerical observer as follows [5]. Using a perceptual image quality metric, JNDmetrix, a lesion image (signal + background) is compared to the corresponding healthy image (background only). Lesion detectability is assumed to be correlated to the metric value which is an indication of the perceived difference between the two images. Contrast masking is one of the effects considered in the derivation of perceptual difference.

The perceptual observer proposed in [5] is double-ended. Our current pipeline is single-ended. This is an advantage in our application since unlike a double-ended observer, a single-ended observer does not require having both a version with and without lesion for every image stack.

Our current pipeline is designed for DBT (three dimensional, 2D space and 1D time; may be used for other 3D modalities, though not tested) and is able, for example, to show a peak in detection performance with slice browsing speed [2]. Numerical observers in above mentioned papers are spatial-only (2D).

## 2 Methods

### 2.1 Simulation platforms and preparation of datasets

In this work, synthetic breast images were generated using the breast anatomy and imaging simulation pipeline developed at the University of Pennsylvania (UPenn). Normal breast anatomy is simulated by a recursive partitioning algorithm using oc-trees [10]. Phantom deformation due to clinical breast positioning and compression is simulated using a finite element model [11]. DBT image acquisition is simulated by ray tracing projections through the phantoms, assuming a polyenergetic x-ray beam without scatter, and an ideal detector model. Reconstructed breast images are obtained using the Real-Time Tomography image reconstruction and processing method [12].

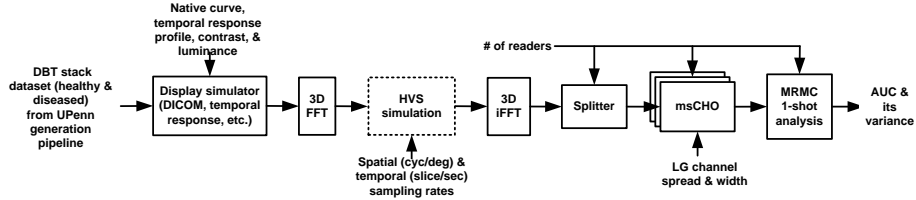

**Fig. 1.** Block diagram of the display and virtual observer simulation. The methods proposed in this paper are used in the dotted block.

The display and virtual observer simulation (Fig. 1) is implemented in MEVIC (Medical Virtual Imaging Chain) [13], an extensible C++ platform developed for medical image processing and visualization at Barco. DBT stack datasets (volumes of interest) with and without simulated lesions, generated using the UPenn pipeline, are input to the display and virtual observer simulation pipeline. For the experiments with numerical observer that are reported here, the "simple background" dataset (see next paragraph for details) consists of 3296 reconstructed 64x64x32 DBT image stacks, half with lesions and half without. Each stack is first decomposed into its spatiotemporal frequency components using a 3D fast Fourier transform (FFT). The stCSF [1, 2], contrast masking (Section 2.2), and psychometric function [1, 2] are modeled in the dotted block in Fig. 1 to determine the perceived amplitude of each frequency component. Then, an inverse 3D FFT is applied to the perceived amplitudes to transform the perceived stack into the space-time domain. Finally, the results are fed to a multi-slice channelized Hotelling observer (msCHO) developed by Platiša *et al* [14]. For further details of the simulation see [2].

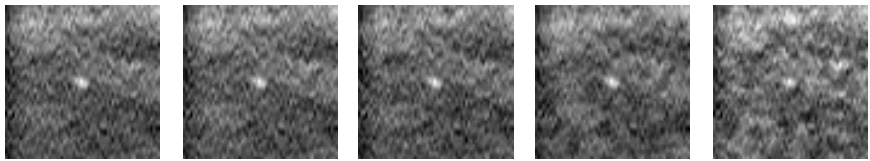

**Fig. 2.** From left to right: Slices #16 from the same image stack in datasets with background complexity level of 0 (simple background), 1, 2, 3, and 4. The insertion contrast for these images is increased to make the lesion (the bright spot in the center of each slice) visible for the purposes of publication.

To create datasets with varying background complexity, spatiotemporal low-pass Gaussian noise with four different levels of energy was added to the dataset from

UPenn phantom. Slices #16 from a sample image stack in five datasets are shown in Fig. 2.

### 2.2 Experiments with numerical observers

The following section describes how we modeled HVS contrast masking property in the numerical observer. According to Winkler, we can disregard temporal contrast masking effect since there is no abrupt change in average luminance (Section 9.2 of [6]). This does assume a continuous browsing in viewing DBT stacks.

To account for spatial contrast masking, we use Barten's model (Chapter 6 of [7]), with the following considerations. Barten addressed the masking of a single spatial tone (single-frequency signal) by a band-pass noise. We assume each frequency component as masker for all other frequency components in a neighborhood determined by masker-maskee difference in frequency and orientation. This is done for every slice in the image stack and the result is used to adjust the CSF-only visibility threshold calculated for each spatio-temporal component of the image stack as follows (Eq. 2.50 in [7]):

$$m'_t = \sqrt{m_t^2 + k^2 m_n^2}, \qquad (1)$$

where $m'_t$ is the visibility threshold with masking, $m_t$ is the CSF-only visibility threshold, $m_n$ is the masker power, and $k$ is Crozier coefficient. The rest of processing is the same as the CSF-only pipeline with psychometric non-linearity described in our prior work [1, 2].

To find $m_n(u, v)$, the masker power for a component with spatial frequency $(u, v)$, we first approximate $S$, the spatial spectrum of the image stack, as follows.

$$S(u, v) = \sum_{\forall w} |I(u, v, w)|^2 \qquad (2)$$

$I$ is 3D DFT of the image stack. $w$ denotes the temporal frequency. By generalizing Eq. 6.2 of [7] for 2D spatial frequencies, we derive the following formula for the masker power.

$$m_n(u, v) = \frac{\sum_{\forall (u', v')} w(u, v, u', v') S(u', v')}{S(0,0) \sum_{\forall (u', v')} w(u, v, u', v')}, \quad u \neq 0 \text{ or } v \neq 0, \qquad (3)$$

$$\text{and } m_n(0,0) = 0.$$

The function $w(u, v, u', v')$ allows a higher weight for nearby components in (spatial) frequency and orientation and is given by

$$w(u,v,u',v') = \begin{cases} 0, & |\alpha| > 5° \\ e^{-2.2ln^2\left(1+\sqrt{\frac{(u-u')^2+(v-v')^2}{u^2+v^2}}\right)}, & |\alpha| \leq 5° \end{cases} \quad (4)$$

where $\alpha$ is the angle between $(u',v')$ and $(u,v)$. This weighting function above is derived by generalizing Eq. 6.4 of [7] for 2D spatial frequencies, and considering the fact that only excitations close in orientation mask one another (Eq. 5.18 in [8]).

Simulation parameters are the same as those used for human observer experiments (Section 2.3). For calculation of stCSF and the perceived amplitudes, see [2].

## 2.3 Experiments with human observers

The following section describes how we have established human observer detection performance in three levels of background complexity. From the same datasets used for experiments with numerical observer, 35 image stacks are randomly chosen for each of the six conditions. The set of conditions is the Cartesian product of {lesion, healthy} (i.e., lesion present or absent) and {0, 2, 4} level of background complexity (see Fig. 2), simulated by the image stacks without addition of noise, and the image stacks with medium and high energy noise added (the noise spectrum remained the same) respectively.

The total of 210 image stacks were presented to the human observer in a random order on a DICOM-calibrated BARCO MDMG-5221 medical display which is optimized and cleared by FDA for reading of DBT images and is equipped with Rapid-Frame temporal response compensation technology. Each image stack was displayed in cine mode at a constant browsing speed of 10 slice/sec twice. The recommended viewing distance from the display was about 40 cm (translates to a spatial sample rate of 18 pixel/degree) but was not strictly enforced for the observer's comfort. The maximum luminance of display ($L_{max}$) was set to 850 cd/m². In our viewing environment, the black point luminance of the display (i.e., the level of luminance associated with drive level of zero) was measured at $L_{min}$ = 1.75 cd/m², using a Minolta CS-100A. Therefore, the effective contrast given by $L_{max}/L_{min}$, was 486.

The observer could repeat the presentation of an image stack (as described above) as many times as desired, or score the presence of a lesion in the spatiotemporal center of the stack. No temporal or spatial clue was provided for the location of lesion. That was because, given our presentation scenario, we found such clues unnecessary and even distracting from the lesion detection task in our pilot experiments. Scoring an image stack consisted of entering one number from the set {0, 1, 2, 3} meaning {certainly no lesion, probably no lesion, probably lesion, certainly lesion} respectively. When an image stack was scored, the process above was repeated for the next image stack. We considered the detection performance as the percentage of correctly identified (i.e., scored 2 or 3) lesion image stacks.

To have more stable results, the same set of image stacks that were randomly chosen for the first human observer was used for the experiments with other human observers as well.

The observers were required to have normal vision and pass a 10-minute training session to become familiar with the experiment and their task. The observers were not radiologist. This is justified considering the fact that the location of the lesion, if present, is always known (and constant), hence the detection task is simply reduced to recognizing a bright spot in a background with various levels of complexity.

## 3    Results and discussion

The results of experiments with two human observers are given in Table 1. As expected, the detection performance for human observers falls with increasing background complexity. It is conceivable that with more experience, human observers who are aware of lesion prevalence rate (50% in our experiments) reach the chance performance (0.5) even in high background complexity.

Table 1. Percentage of correctly identified lesion stacks (of total) in three background complexity levels.

| Background Complexity | Low | Medium | High |
|---|---|---|---|
| Observer **A** | 0. 9714 | 0. 7714 | 0. 2571 |
| Observer **B** | 0. 8000 | 0. 7143 | 0. 3429 |

Simulation results from modeling HVS with stCSF only, and stCSF plus contrast masking are compared in Fig. 3. In each case, three methods of calculating the perceived amplitudes using the visibility threshold are simulated: Monte Carlo (MC), probability map (PM), and linear filtering (LF). While both methods PM & MC use a nonlinear psychometric function, PM is deterministic and MC is not. For more information on calculation of the perceived amplitudes, see [2].

The inclusion of contrast masking makes the numerical observer more closely resemble the human observer performance, in the sense that it removes some of the over performance of numerical observer. Among the six graphs in Fig. 3, the one showing the results of modeling HVS with stCSF plus contrast masking using the PM method for perceived amplitude calculation demonstrates the most significant drop in detection performance with increasing background complexity. Even this graph, however, cannot match the fast drop of detection performance of human observers as listed in Table 1.

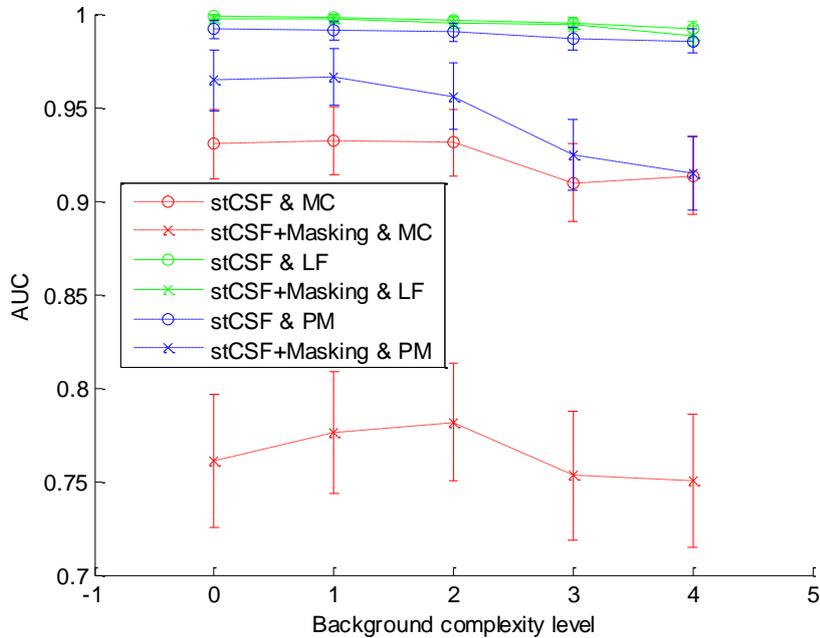

**Fig. 3.** Detection performance (in AUC) for datasets at various background complexity levels (Section 2.1).

## 4 Conclusion

Our results indicate that by modeling the HVS contrast masking property, lesion detection becomes more difficult in a busier background, as expected from a typical human observer.

This is a work in progress. We plan to continue our work on this project in the following avenues. (i) Experiments are conducted with more human observers and the scores will be aggregated with one-shot multi-reader multi-case ROC analysis [9]. (ii) A more realistic model of background complexity including anatomical and quantum noise using our software phantom will be used to prepare the datasets. (iii) We will explore the extent of our claims for a variety of lesion sizes and types.

## Acknowledgement

This work is supported by the US National Institutes of Health (grant 1R01CA154444). Ali Avanaki would like to thank Drs. Kyle Myers, Cédric Marchessoux, and Miguel Eckstein. Dr. Maidment is on the scientific advisory board of Real-Time Tomography, LLC.

# References

[1] A.N. Avanaki, K.S. Espig, C. Marchessoux, E.A. Krupinski, P.R. Bakic, T.R.L. Kimpe, A.D.A. Maidment, "On modeling the effects of display contrast and luminance in a spatio-temporal numerical observer," *MIPS* presentation, Washington DC, 2013.

[2] A.N. Avanaki, K.S. Espig, A.D.A. Maidment, C. Marchessoux, P.R. Bakic, T.R.L. Kimpe, "Development and evaluation of a 3D model observer with nonlinear spatiotemporal contrast sensitivity," to appear in *Proc. of SPIE Medical Imaging,* 2014.

[3] A.N. Avanaki, K.S. Espig, C. Marchessoux, E.A. Krupinski, P.R. Bakic, T.R.L. Kimpe, A.D.A. Maidment, "Integration of spatio-temporal contrast sensitivity with a multi-slice channelized Hotelling observer," *Proc. SPIE Medical Imaging*, 2013.

[4] Y. Zhang, B.T. Pham, M.P. Eckstein, "The Effect of Nonlinear Human Visual System Components on Performance of a Channelized Hotelling Observer Model in Structured Backgrounds," *IEEE Trans. on Medical Imaging*, vol. 25, pp. 1348-1362, 2006.

[5] E. A. Krupinski, J. Lubin, H. Roehrig, J. Johnson, J. Nafziger "Using a human visual system model to optimize soft-copy mammography display: influence of veiling glare," *Acad Radiol.*, vol. 13, pp. 289-295, 2006.

[6] S. Winkler, "Issues in vision modeling for perceptual video quality assessment," *Signal Processing*, vol. 78, pp. 231-252, 1999.

[7] P.G.J. Barten, *Contrast sensitivity of the human eye and its effects on image quality*, SPIE Optical Engineering Press, Bellingham, WA, 1999.

[8] M. Barni, F. Bartolini, *Watermarking Systems Engineering: Enabling Digital Assets Security and Other Applications*, CRC Press, 2004.

[9] B.D. Gallas, "One-shot estimate of MRMC variance: AUC," *Acad Radiol.*, vol. 13, pp. 353-362, 2006.

[10] D. Pokrajac, A.D.A. Maidment, P.R. Bakic, "Optimized Generation of High Resolution Breast Anthropomorphic Software Phantoms," *Medical Physics*, vol. 39, pp. 2290-2302, April 2012.

[11] M.A. Lago, A.D.A. Maidment, P.R. Bakic, "Modelling of mammographic compression of anthropomorphic software breast phantom using FEBio," *Proc. Int'l Symposium on Computer Methods in Biomechanics and Biomedical Engineering* (CMBBE) Salt Lake City, UT, 2013.

[12] J. Kuo, P. Ringer, S.G. Fallows, S. Ng, P.R. Bakic, A.D.A. Maidment, "Dynamic reconstruction and rendering of 3D tomosynthesis images" *Proc. of SPIE,* Medical Imaging 2011.

[13] C. Marchessoux, T. R. L. Kimpe, and T. Bert, "A virtual image chain for perceived and clinical image quality of medical display," *J. of Display Technology*, vol. 4, pp. 356–368, 2008.

[14] L. Platiša, B. Goossens, E. Vansteenkiste, S. Park, B. Gallas, A. Badano and W. Philips, "Channelized hotelling observers for the assessment of volumetric imaging data sets," *J. of Optical Society of America A*, vol. 28, pp. 1145 – 1163, 2011.